\documentclass[runningheads]{llncs}
\usepackage{amsmath,amssymb} 
\usepackage{color}
\usepackage{epsfig}
\usepackage{graphicx}
\usepackage{amsmath}
\usepackage{amssymb}
\usepackage{enumitem}
\usepackage[normalem]{ulem}
\usepackage[ruled]{algorithm2e}
\usepackage{multirow}
\usepackage{array}
\usepackage{ctable} 
\usepackage{makecell}
\usepackage[capposition=bottom]{floatrow}
\usepackage{comment}
\DeclareMathAlphabet{\mathcal}{OMS}{cmsy}{m}{n}
\usepackage{textcomp}
\usepackage{dblfloatfix}
\usepackage{caption}
\usepackage{subcaption}
\usepackage{colortbl,xcolor}
\usepackage{booktabs}
\usepackage{mathtools}
\usepackage[pagebackref=false,breaklinks=true,letterpaper=true,colorlinks,citecolor=blue,linkcolor=blue,bookmarks=false]{hyperref}
\usepackage{textpos}

\newcommand{\name}{\textsc{VisualEchoes}}

\definecolor{customgray}{rgb}{0.9, 0.9, 0.9}
\newcolumntype{g}{>{\columncolor{customgray}}c}
\newcolumntype{z}{>{\columncolor{customgray}}l}
\newcolumntype{?}[1]{!{\vrule width #1}}
\renewcommand{\paragraph}[1]{{\vspace{2mm}\noindent\textbf{#1}\,\,}}

\begin{document}
\pagestyle{headings}
\mainmatter
\def\ECCVSubNumber{927}  

\title{\name: Spatial Image Representation\\Learning through Echolocation} 

\titlerunning{\name}
%
%
\author{Ruohan Gao\textsuperscript{1,3}, Changan Chen\textsuperscript{1,3}, Ziad Al-Halah\textsuperscript{1}, \\Carl Schissler\textsuperscript{2}, Kristen Grauman\textsuperscript{1,3}  }

\authorrunning{Gao et al.}
%
\institute{The University of Texas at Austin \and Facebook Reality Lab \and Facebook AI Research}

\institute{\textsuperscript{1}The University of Texas at Austin, \textsuperscript{2}Facebook Reality Lab, \textsuperscript{3}Facebook AI Research\\
\email{\{rhgao,changan,ziad,grauman\}@cs.utexas.edu}, \email{carl.schissler@fb.com}}


\maketitle

\begin{abstract}
Several animal species (e.g., bats, dolphins, and whales) and even visually impaired humans have the remarkable ability to perform echolocation: a biological sonar used to perceive spatial layout and locate objects in the world. We explore the spatial cues contained in echoes and how they can benefit vision tasks that require spatial reasoning. First we capture echo responses in photo-realistic 3D indoor scene environments. Then we propose a novel interaction-based representation learning framework that learns useful \emph{visual} features via echolocation. We show that the learned image features are useful for multiple downstream vision tasks requiring spatial reasoning---monocular depth estimation, surface normal estimation, and visual navigation---with results comparable or even better than heavily supervised pre-training. Our work opens a new path for representation learning for embodied agents, where supervision comes from interacting with the physical world.
\end{abstract}

\begin{textblock*}{\textwidth}(0cm,-13cm)
\centering
In Proceedings of the European Conference on Computer Vision (ECCV), 2020%
\end{textblock*}


\section{Introduction}~\label{sec:intro}
\vspace{-0.2in}

The perceptual and cognitive abilities of embodied agents are inextricably tied to their physical being. We perceive and act in the world by making use of all our senses---especially looking and listening.  We see our surroundings to avoid obstacles, listen to the running water tap to navigate to the kitchen, and infer how far away the bus is once we hear it approaching.

By using \emph{two} ears, we perceive spatial sound.  Not only can we identify the sound-emitting object (e.g., the revving engine corresponds to a bus), but also we can determine that object's location, based on the time difference between when the sound reaches each ear (Interaural Time Difference, ITD) and the difference in sound level as it enters each ear (Interaural Level Difference, ILD). Critically, even beyond objects, audio is also rich with information about the \emph{environment} itself.  The sounds we receive are a function of the geometric structure of the space around us and the materials of its major surfaces~\cite{antonacci2012inference}. In fact, some animals capitalize on these cues by using \emph{echolocation}---actively emitting sounds to perceive the 3D spatial layout of their surroundings~\cite{rosenblum2000echolocating}.

We propose to learn image representations from echoes. Motivated by how animals and blind people obtain spatial information from echo responses, first we explore to what extent the echoes of chirps generated in a scanned 3D environment are predictive of the depth in the scene. Then, we introduce \name, a novel image representation learning method based on echolocation. Given a first-person RGB view and an echo audio waveform, our model is trained to predict the correct camera orientation at which the agent would receive those echoes. In this way, the representation is forced to capture the alignment between the sound reflections and the (visually observed) surfaces in the environment.  At test time, we observe only pixels---no audio.  Our learned \name~encoder better reveals the 3D spatial cues embedded in the pixels, as we demonstrate in three downstream tasks.

Our approach offers a new way to learn image representations without manual supervision by \emph{interacting} with the environment.  In pursuit of this high-level goal there is exciting---though limited---prior work that learns visual features by touching objects~\cite{owens2016visually,purushwalkam2019bounce,agrawal2016learning,pinto-icra2016} or moving in a space~\cite{jayaraman2015learning,agrawal2015learning,gandhi2017learning}. Unlike mainstream ``self-supervised" feature learning work that crafts pretext tasks for large static repositories of human-taken images or video (e.g., colorization~\cite{zhang2016colorful}, jigsaw puzzles~\cite{noroozi2016unsupervised}, audio-visual correspondence~\cite{Korbar2018cotraining,arandjelovic2017look}), in \emph{interaction-based feature learning} an embodied agent\footnote{person, robot, or simulated robot} performs physical actions in the world that dynamically influence its own first-person observations and possibly the environment itself.  Both paths have certain advantages:  while conventional self-supervised learning can capitalize on massive static datasets of human-taken photos, interaction-based learning
allows an agent to ``learn by acting" with rich multi-modal sensing. This has the advantage of learning features adaptable to new environments. Unlike any prior work, we explore feature learning from echoes.

Our contributions are threefold: 1) We explore the spatial cues contained in echoes, analyzing how they inform depth prediction; 2) We propose \name, a novel interaction-based feature learning framework that uses echoes to learn an image representation and does not require audio at test time; 3) We successfully validate the learned spatial representation for the fundamental downstream vision tasks of monocular depth prediction, surface normal estimation, and visual navigation, with results comparable to or even outperforming heavily supervised pre-training baselines.

\vspace*{-0.15in}
\section{Related Work}~\label{sec:related}
\vspace*{-0.35in}

\paragraph{Auditory Scene Analysis using Echoes}
Previous work shows that using echo responses only, one can predict 2D~\cite{antonacci2012inference} or 3D~\cite{dokmanic2013acoustic} room geometry and object shape~\cite{Frank2020ComparingVT}. Additionally, echoes can complement vision, especially when vision-based depth estimates are not reliable, e.g., on transparent windows or featureless walls~\cite{kim20173d,ye20153d}. In dynamic environments, autonomous robots can leverage echoes for obstacle avoidance~\cite{vanderelst2015sensorimotor} or mapping and navigation~\cite{eliakim2018fully} using a bat-like echolocation model. Concurrently with our work, a low-cost audio system called BatVision is used to predict depth maps purely from echo responses~\cite{batvision}. Our work explores a novel direction for auditory scene analysis by employing echoes for spatial visual feature learning, and unlike prior work, the resulting features are applicable in the absence of any audio.

\vspace*{-0.1in}

\paragraph{Self-Supervised Image Representation Learning}
Self-supervised image feature learning methods leverage structured information within the data itself to generate labels for representation learning~\cite{de1994learning,goyal2019scaling}. To this end, many ``pretext" tasks have been explored---for example, predicting the rotation applied to an input image~\cite{gidaris2018unsupervised,agrawal2015learning}, discriminating image instances~\cite{feng2019self}, colorizing images~\cite{larsson2017colorization,zhang2016colorful}, solving a jigsaw puzzle from image patches~\cite{noroozi2016unsupervised}, predicting unseen views of 3D objects~\cite{jayaraman2018shapecodes}, or multi-task learning using synthetic imagery~\cite{ren2018cross}. Temporal information in videos also permits self-supervised tasks, for example, by predicting whether a frame sequence is in the correct order~\cite{misra2016shuffle,fernando2017self} or ensuring visual coherence of tracked objects~\cite{wang2015unsupervised,gao2016object-centric,slow-steady}.
Whereas these methods aim to learn features generically useful for recognition, our objective is to learn features generically useful for spatial estimation tasks. Accordingly, our echolocation objective is well-aligned with our target family of spatial tasks (depth, surfaces, navigation), consistent with findings that task similarity is important for positive transfer~\cite{Zamir_2018_CVPR}. Furthermore, unlike any of the above, rather than learn from massive repositories of human-taken photos, the proposed approach learns from interactions with the scene via echolocation.

\vspace*{-0.1in}

\paragraph{Feature Learning by Interaction} Limited prior work explores feature learning through interaction.  Unlike the self-supervised methods discussed above, this line of work fosters agents that learn from their own observations in the world, which can be critical for adapting to new environments and to realize truly ``bottom-up" learning by experience.  Existing methods explore touch and motion interactions. In~\cite{owens2016visually}, objects are struck with a drumstick to facilitate learning material properties when they sound.  In~\cite{purushwalkam2019bounce}, the trajectory of a ball bouncing off surfaces facilitates learning physical scene properties. 
In~\cite{pinto-icra2016,agrawal2016learning}, a robot learns object properties by poking or grasping at objects.  In~\cite{gandhi2017learning}, a drone learns not to crash after attempting many crashes. In~\cite{jayaraman2015learning,agrawal2015learning}, an agent tracks its egomotion in concert with its visual stream to facilitate learning visual categories. In contrast, our idea is to learn visual features by \emph{emitting audio} to acoustically interact with the scene. Our work offers a new perspective on interaction-based feature learning and has the advantages of not disrupting the scene physically and being ubiquitously available, i.e., reaching all surrounding surfaces.

\vspace*{-0.1in}

\paragraph{Audio-Visual Learning}
Inspiring recent work integrates sound and vision in joint learning frameworks that synthesize sounds for video~\cite{owens2016visually,zhou2017visual}, spatialize monaural sounds from video~\cite{gao2019visualsound,morgadoNIPS18}, separate sound sources~\cite{owens2018audio,ephrat2018looking,gao2018objectSounds,zhao2018sound,gao2019coseparation,gan2020music}, perform cross-modal feature learning~\cite{aytar2016soundnet,owens2016ambient}, track audio-visual targets~\cite{gebru2015iccvw,ban2018icassp,alameda2015salsa,gan2019tracking}, segment objects with multi-channel audio~\cite{irie2019seeing}, direct embodied agents to navigate in indoor environments~\cite{changan-eccv2020,gan2020look}, recognize actions in videos~\cite{gao2020listentolook,kazakos2019TBN}, and localize pixels associated with sounds in video frames~\cite{tian2018audio,Senocak_2018_CVPR,arandjelovic2017objects,hershey2000audio}. None of the prior methods pursues echoes for visual learning.  Furthermore, whereas nearly all existing audio-visual methods operate in a passive manner, observing incidental sounds within a video, in our approach the system learns by actively emitting sound---a form of interaction with the physical environment.

\paragraph{Monocular Depth Estimation}
To improve monocular depth estimation, recent methods focus on improving neural network architectures~\cite{fu2018deep} or graphical models~\cite{wang2015towards,liu2015deep,xu2017multi}, employing multi-scale feature fusion and multi-task learning~\cite{eigen2015predicting,hu2019revisiting}, leveraging motion cues from successive frames~\cite{Ummenhofer_2017_CVPR}, or transfer learning~\cite{karsch2014depth}. However, these approaches rely on depth-labeled data that can be expensive to obtain. Hence, recent approaches leverage scenes' spatial and temporal structure to self-supervise depth estimation, by using the camera motion between pairs of images~\cite{garg2016unsupervised,godard2017unsupervised} or frames~\cite{zhou2017unsupervised,vijayanarasimhan2017sfm,godard2019digging,jiang2018self}, or consistency cues between depth and features like surface normals~\cite{yang2018unsupervised} or optical flow~\cite{Ranjan_2019_CVPR}. Unlike any of these existing methods, we show that audio in the form of an echo response can be effectively used to recover depth, and we develop a novel feature learning method that benefits a purely visual representation (no audio) at test time.
\vspace*{-0.1in}
\section{Approach}~\label{sec:approach}
\vspace*{-0.2in}

Our goals are to show that echoes convey spatial information, to learn visual representations by echolocation, and to leverage the learned representations for downstream tasks. In the following, we first describe how we simulate echoes in 3D environments (Sec.~\ref{sec:simulation}). Then we perform a case study to demonstrate how echoes can benefit monocular depth prediction (Sec.~\ref{sec:case-study}). Next, we present \name, our interaction-based feature learning formulation to learn image representations (Sec.~\ref{sec:feature_learning}).
Finally, we exploit the learned visual representation for monocular depth, surface normal prediction, and visual navigation (Sec.~\ref{sec:downstream_tasks}).

\vspace*{-0.15in}
\subsection{Echolocation Simulation}~\label{sec:simulation}
\vspace*{-0.2in}

Our echolocation simulation is based on recent work on audio-visual navigation \cite{changan-eccv2020}, which builds a realistic acoustic simulation on top of the Habitat \cite{savva_habitat:_2019} platform and Replica environments \cite{straub2019replica}. Habitat~\cite{savva_habitat:_2019} is an open-source 3D simulator that supports efficient RGB, depth, and semantic rendering for multiple datasets \cite{straub2019replica,chang_matterport3d_2017,xia_gibson_2018}. Replica is a dataset of 18 apartment, hotel, ofﬁce, and room scenes with 3D meshes and high deﬁnition range (HDR) textures and renderable reﬂector information. The platform in~\cite{changan-eccv2020} simulates acoustics by pre-computing room impulse responses (RIR) between all pairs of possible source and receiver locations, using a form of audio ray-tracing~\cite{veach1995bidirectional}.  An RIR is a transfer function between the sound source and the sound microphone, and it is influenced by the room geometry, materials, and the sound source location \cite{room_acoustics_taylor}. The sound received at the listener location is computed by convolving the appropriate RIR with the waveform of the source sound.

We use the binaural RIRs for all Replica environments to generate echoes for our approach. As the source audio ``chirp" we use a sweep signal from 20Hz-20kHz (the human-audible range) within a duration of 3ms. While technically any emitted sound could provide some echo signal from which to learn, our design (1) intentionally provides the response for a wide range of frequencies and (2) does so in a short period of time to avoid overlap between echoes and direct sounds. We place the source at the \emph{same} location as the receiver and convolve the RIR for this source-receiver pair with the sweep signal.  In this way, we compute the echo responses that would be received at the agent's microphone locations. We place the agents at all navigable points on the grid (every 0.5m~\cite{changan-eccv2020}) and orient the agent in  four cardinal directions ($0^{\circ}$, $90^{\circ}$, $180^{\circ}$, $270^{\circ}$) so that the rendered egocentric views (RGB and depth) and echoes capture room geometry from different locations and orientations.

\begin{figure}[t]
    \center
    \includegraphics[width=\linewidth]{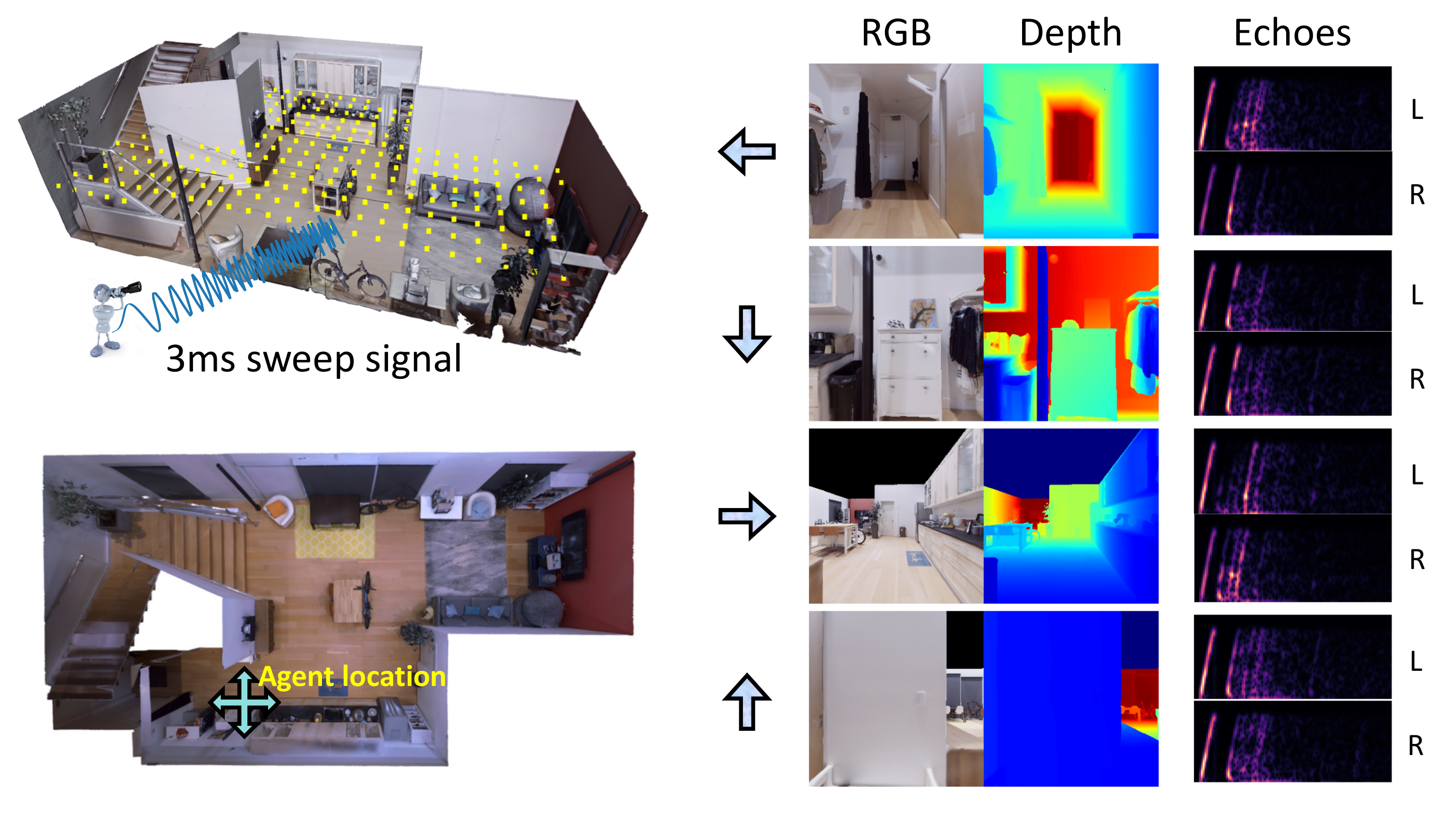}
    \caption{Echolocation simulation in real-world scanned environments. During training, the agent goes to the densely sampled locations marked with yellow dots. The left bottom figure illustrates the top-down view of one Replica scene where the agent's location is marked. The agent  actively emits 3 ms omnidirectional sweep signals to get echo responses from the room. The right column shows the corresponding RGB and depth of the agent's view as well as the echoes received in the left and right ears when the agent faces each of the four directions.}
    \label{fig:echolocation_illustration}
    \vspace*{-0.1in}
\end{figure}

Fig.~\ref{fig:echolocation_illustration} illustrates how we perform echolocation for one scene environment. The agent goes to the densely sampled navigable locations marked with yellow dots and faces four orientations at each location. It actively emits omnidirectional chirp signals and records the echo responses received when facing each direction. Note that the spectrograms of the sounds received at the left (L) and right (R) ears reveal that the agent first receives the direct sound (strong bright curves), and then receives different echoes for the left and right microphones due to ITD, ILD, and pinnae reflections. The subtle difference in the two spectrograms conveys cues about the spatial configuration of the environment, as can be observed in the last column of Fig.~\ref{fig:echolocation_illustration}.


\begin{figure}[t]
    \center
    \includegraphics[width=0.95\linewidth]{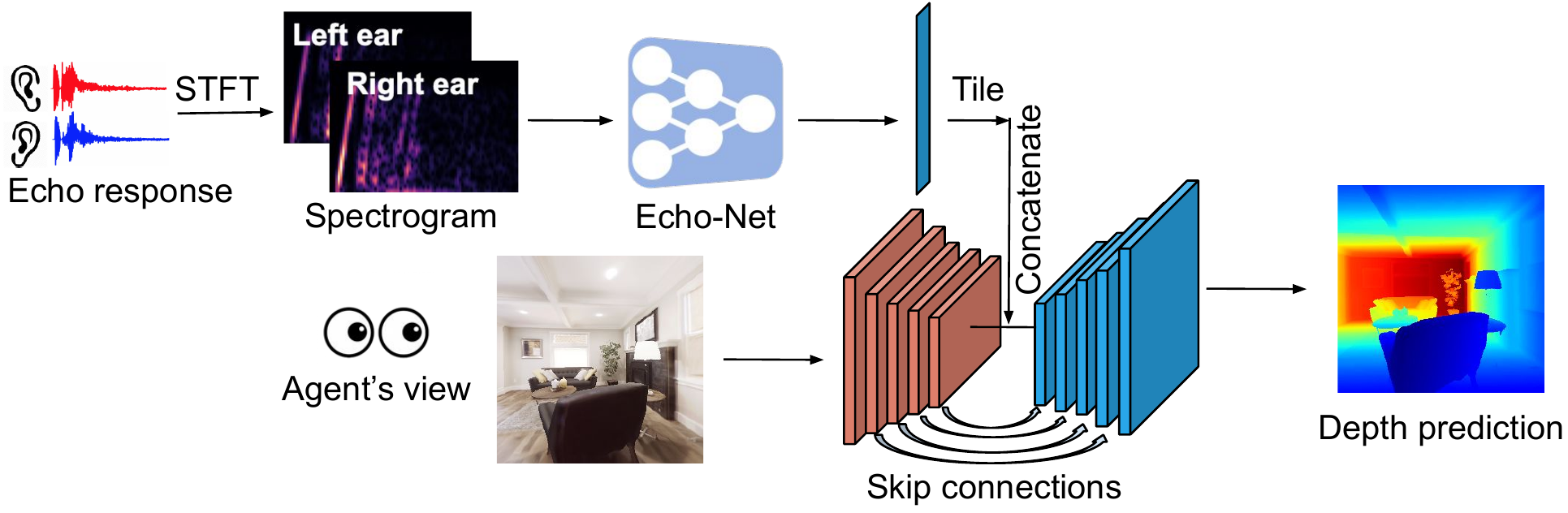}
    \caption{Our \textsc{RGB+Echo2Depth} network takes the echo responses and the corresponding egocentric RGB view as input, and performs joint audio-visual analysis to predict the depth map for the input image. The injected echo response provides additional cues of the spatial layout of the scene. Note: in later sections we define networks that do not have access to the audio stream at test time.}
    \label{fig:rgbecho2depth}
    \vspace*{-0.1in}
\end{figure}

\vspace*{-0.15in}
\subsection{Case Study: Spatial Cues in Echoes}~\label{sec:case-study}
\vspace*{-0.2in}

With the synchronized egocentric views and echo responses in hand, we now conduct a case study to investigate the spatial cues contained in echo responses in these realistic indoor 3D environments. We have two questions: (1) can we directly predict depth maps purely from echoes? and (2) can we use echoes to augment monocular depth estimation from RGB? Answering these questions will inform our ultimate goal of devising a interaction-supervised visual feature learning approach leveraging echoes only at training time (Sec.~\ref{sec:feature_learning}).  Furthermore, it can shed light on the extent to which low-cost audio sensors can replace depth sensors, which would be especially useful for navigation robots under severe bandwidth or sensing constraints, e.g., nano drones~\cite{palossi201964,mcguire2019minimal}. 

Note that these two goals are orthogonal to that of prior work performing depth prediction from a single view~\cite{eigen2014depth,liu2015deep,xu2017multi,fu2018deep,hu2019revisiting}.  Whereas they focus on developing sophisticated loss functions and architectures, here we explore how an agent \emph{actively interacting with the scene acoustically} may improve its depth predictions.  Our findings can thus complement existing monocular depth models.

We devise an \textsc{RGB+Echo2Depth} network (and its simplified variants using only RGB or echo) to test the settings of interest. The \textsc{RGB+Echo2Depth} network predicts a depth map based on the agent's egocentric RGB input and the echo response it receives when it emits a chirp standing at that position and orientation in the 3D environment. The core model is a multi-modal U-Net~\cite{ronneberger2015u}; see Fig.~\ref{fig:rgbecho2depth}. To directly measure the spatial cues contained in echoes alone, we also test a variant called \textsc{Echo2Depth}. Instead of performing upsampling based on the audio-visual representation, this model drops the RGB input, reshapes the audio feature, and directly upsamples from the audio representation. Similarly, to measure the cues contained in the RGB alone, a variant called  \textsc{RGB2Depth} drops the echoes and predicts the depth map purely based on the visual features. The \textsc{RGB2Depth} model represents existing monocular depth prediction approaches that predict depth from a single RGB image, in the context of the same architecture design as \textsc{RGB+Echo2Depth} to allow apples-to-apples calibration of our findings. We use RGB images of spatial dimension $128 \times 128$. See Supp.~for network details and loss functions used to train the three models.

\begin{figure*}[t!]
    \center
    \includegraphics[width=0.9\linewidth]{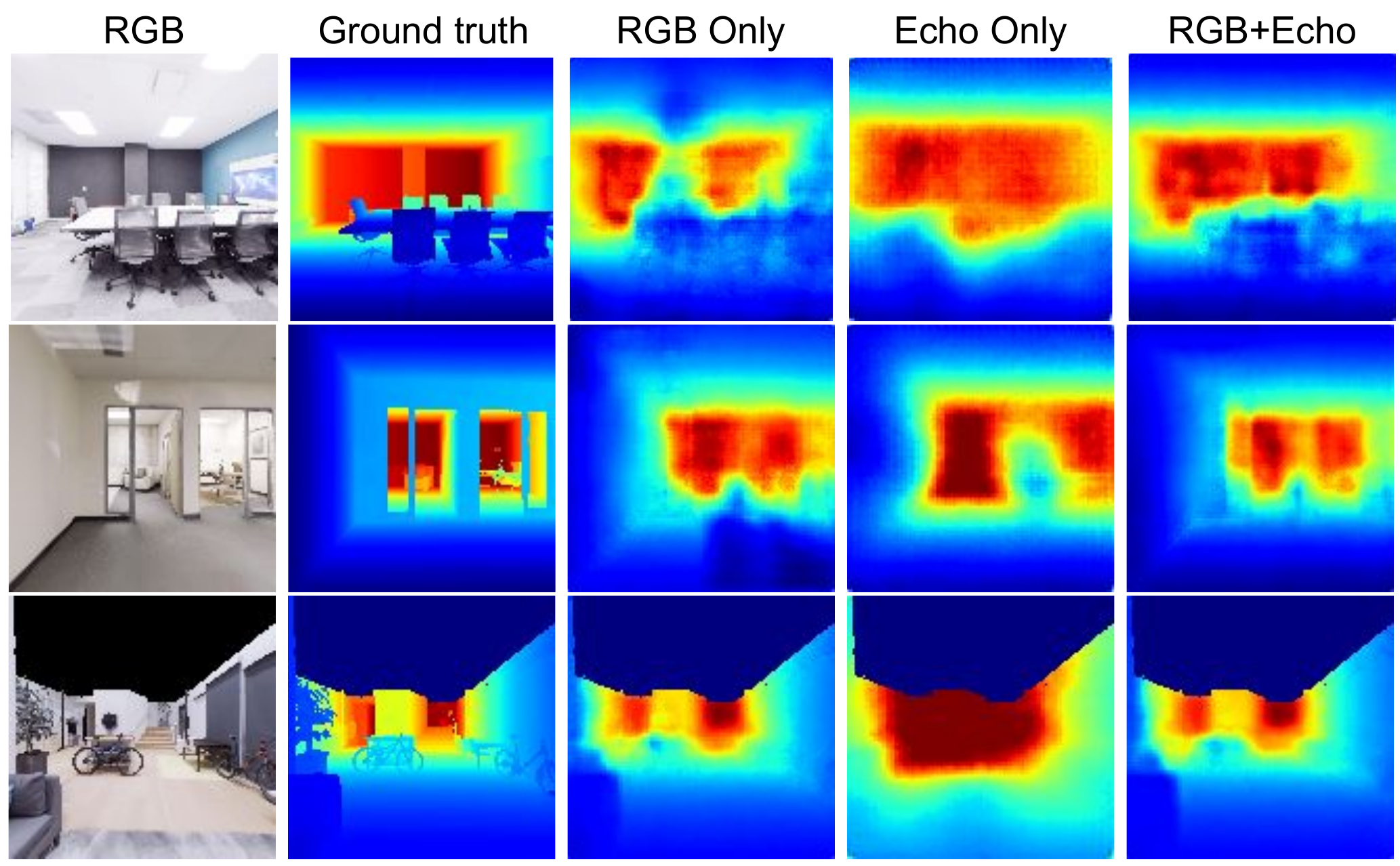}
    \caption{Qualitative results of our case study on monocular depth estimation in unseen environments using echoes. Together with the quantitative results (Tab.~\ref{table:case_study}), these examples show that echoes contain useful spatial cues that  inform a visual spatial task. For example, in row 1, the RGB+Echo model better infers the depth of the column on the back wall, whereas the RGB-Only model mistakenly infers the strong contours to indicate a much closer surface. The last row shows a typical failure case (see text). See Supp.~for more examples.
    }
    \label{fig:case_study_qualitative}
    \vspace*{-0.05in}
\end{figure*}

\begin{table*}[t!]
\fontsize{7.5}{9} \selectfont
\begin{tabular}{zc|c|c|c|c|c}
\toprule
              & ~~RMS~~~$\downarrow$ & ~~REL~~~$\downarrow$ & $~~\log10~~\downarrow$ & $\delta<1.25~\uparrow$ &  $\delta<1.25^2~\uparrow$ &  $\delta<1.25^3~\uparrow$ \\ \hline
\textsc{Average}       &  1.070   &  0.791   &    0.230    &    0.235            &       0.509         &   0.750             \\ 
\textsc{Echo2Depth}    &   0.713   &  0.347   &    0.134    &    0.580            &    0.772            &     0.868           \\ 
\textsc{RGB2Depth}     &  0.374    &   0.202  &    0.076    &   0.749             &    0.883             &    0.945             \\
\textsc{RGB+Echo2Depth} &  \textbf{0.346}   &  \textbf{0.172}   &    \textbf{0.068}    &   \textbf{0.798}             &      \textbf{0.905}          &    \textbf{0.950}            \\ \bottomrule
\end{tabular}
\caption{Case study depth prediction results. $\downarrow$ lower better, $\uparrow$ higher better.}
\label{table:case_study}
\vspace*{-0.1in}
\end{table*}

Table~\ref{table:case_study} shows the quantitative results of predicting depth from only echoes, only RGB, or their combination. We evaluate on a heldout set of three Replica environments (comprising 1,464 total views) with standard metrics: root mean squared error (RMS), mean relative error (REL), mean $\log10$ error ($\log 10$), and thresholded accuracy~\cite{hu2019revisiting,eigen2014depth}. We can see that depth prediction is possible purely from echoes. Augmenting traditional single-view depth estimation with echoes (bottom row) achieves the best performance by leveraging the additional acoustic spatial cues. Echoes alone are naturally weaker than RGB alone, yet still better than the simple \textsc{Average} baseline that  predicts the average depth values in all training data.

Fig.~\ref{fig:case_study_qualitative} shows qualitative examples. It is clear that echo responses indeed contain cues of the spatial layout; the depth map captures the rough room layout, especially its large surfaces. When combined with RGB, the predictions are more accurate. The last row shows a typical failure case, where the echoes alone cannot capture the depth as well due to far away surfaces with weaker echo signals.

\vspace*{-0.15in}
\subsection{\name~Spatial Representation Learning Framework}~\label{sec:feature_learning}
\vspace*{-0.2in}

Having established the scope for inferring depth from echoes, 
we now present our \name~model to leverage echoes for visual representation learning.  We stress that our approach assumes audio/echoes are available only during training; at test time, an RGB image alone is the input.

The key insight of our approach is that the echoes and visual input should be consistent.  This is because both are functions of the same latent variable---the 3D shape of the environment surrounding the agent's co-located camera and microphones.  We implement this idea by training a network to predict their correct association.

In particular, as described in Sec.~\ref{sec:simulation}, at any position in the scene, we suppose the agent can face four orientations, i.e., at an azimuth angle of $0^{\circ}$, $90^{\circ}$, $180^{\circ}$, and $270^{\circ}$. When the agent emits the sweep signal (chirp) at a certain position, it will hear different echo responses when it faces each different orientation.  If the agent correctly interprets the spatial layout of the current view from \emph{visual} information, it should be able to tell whether that visual input is congruous with the echo response it hears. Furthermore, and more subtly, to the extent the agent implicitly learns about probable views surrounding its current egocentric field of view (e.g., what the view just to its right may look like given the context of what it sees in front of it), it should be able to tell which direction the received echo \emph{would} be congruous with, if not the current view.

\begin{figure*}[t!]
    \center
    \includegraphics[width=\linewidth]{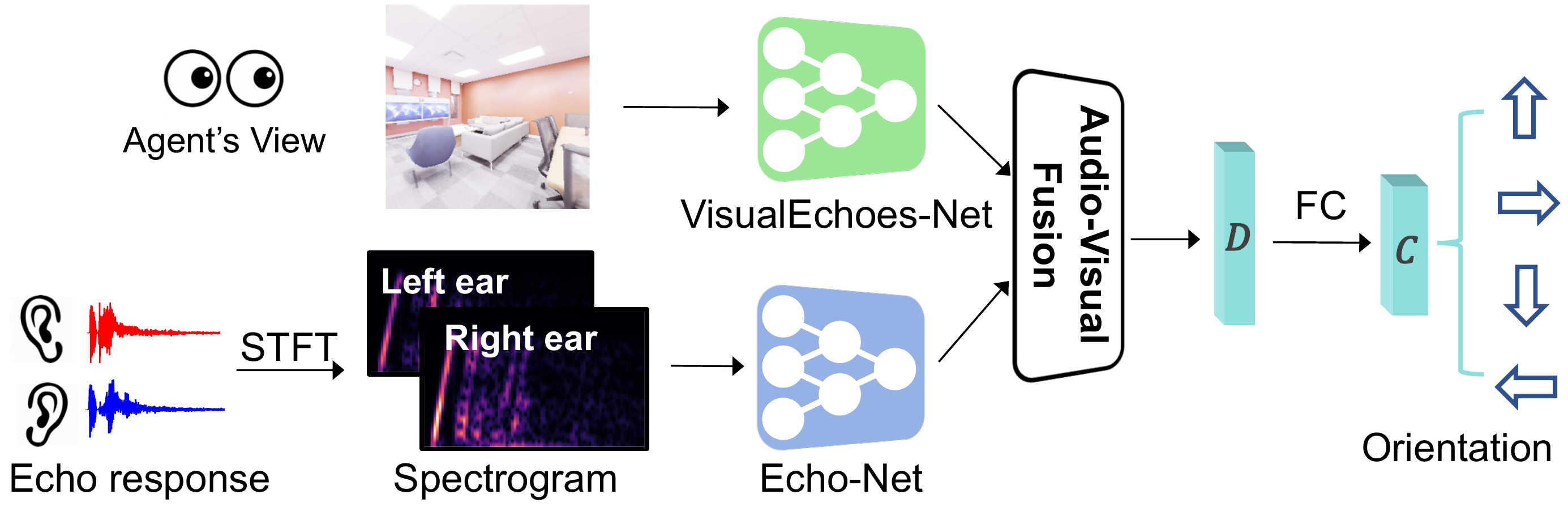}
    \caption{Our \name~network takes the agent's current RGB view as visual input, and the echo responses from one of the four orientations as audio input. The goal is to predict the orientation at which the agent would receive the input echoes based on analyzing the spatial layout in the image. After training with RGB and echoes, the \name-Net is a pre-trained encoder ready to extract spatially enriched features from novel RGB images, as we validate with multiple downstream tasks (cf.~Sec.~\ref{sec:downstream_tasks}).
    }
    \label{fig:network}
\end{figure*}

We introduce a representation learning network to capture this insight. See Fig.~\ref{fig:network}. The visual stream takes the agent's current RGB view as input, and the audio stream takes the echo response received from one of the four orientations---not necessarily the one that coincides with the visual stream orientation. The fusion layer fuses the audio and visual information to generate an audio-visual feature of dimension $D$. A final fully-connected layer is used to make the final prediction among four classes. See Supp.~and Sec.~\ref{sec:results} for architecture details.

The four classes are defined as follows:
\vspace*{-0.1in}
\begin{enumerate}
    \item[$\uparrow$]: The echo is received from the same orientation as the agent's current view.
    \item[$\rightarrow$]: The echo is received from the orientation if the agent turns right by $90^{\circ}$.
    \item[$\downarrow$]: The echo is received from the orientation opposite the agent's current view.
    \item[$\leftarrow$]: The echo is received from the orientation if the agent turns left by $90^{\circ}$.
\end{enumerate}
\vspace*{-0.05in}

The network is trained with cross-entropy loss. Note that although the emitted source signal is always the same (3 ms \emph{omnidirectional} sweep signal, cf.~Sec.~\ref{sec:simulation}), the agent hears different echoes when facing the four directions because of the shape of the ears and the head shadowing effect modeled in the binaural head-related transfer function (HRTF). Since the classes above are defined relative to the agent's current view, it can only tell the orientation for which it is receiving the echoes if it can correctly interpret the 3D spatial layout within the RGB input. In this way,  the agent's aural interaction with the scene enhances  spatial feature learning for the visual stream.

The proposed idea generalizes trivially to use more than four discrete orientations---and even arbitrary orientations if we were to use regression rather than classification.  The choice of four is simply based on the sound simulations available in existing data~\cite{changan-eccv2020}, though we anticipate it is a good granularity to capture the major directions around the agent. Our training paradigm requires the representation to discern mismatches between the image and echo using echoes generated from the same physical position on the ground plane but different orientations.
This is in line with our interactive embodied agent motivation, where an agent can look ahead, then turn and hear echoes from another orientation at the same place in the environment, and learn their (dis)association. 
In fact, ecological psychologists report that humans can perform more accurate echolocation when moving, supporting the rationale of our design~\cite{stroffregen1995human,rosenblum2000echolocating}. Furthermore, our design ensures the mismatches are ``hard" examples useful for learning spatial features because the audio-visual data at offset views will naturally be related to one another (as opposed to views or echoes from an unrelated environment).

\vspace{-0.15in}
\subsection{Downstream Tasks for the Learned Spatial Representation}~\label{sec:downstream_tasks}
\vspace*{-0.2in}

Having introduced our \name~feature learning framework, next we describe how we repurpose the learned visual representation for three fundamental downstream tasks that require spatial reasoning: monocular depth prediction, surface normal estimation, and visual navigation. For each task, we adopt strong models from the literature and swap in our pre-trained encoder \name-Net for the RGB input. 

\vspace*{-0.05in}
\paragraph{Monocular depth prediction:}  We explore how 
our echo-based pre-training can benefit performance for traditional monocular depth prediction.  Note that unlike the case study in Sec.~\ref{sec:case-study}, in this case there are no echo inputs at test time, only RGB. 
To evaluate the quality of our learned representation, we adopt a strong recent approach for monocular depth prediction~\cite{hu2019revisiting} consisting of several novel loss functions and a multi-scale network architecture that is based on a backbone network. We pre-train ResNet-50~\cite{he2016deep} using \name~and use it as the backbone for comparison with~\cite{hu2019revisiting}.

\vspace*{-0.05in}
\paragraph{Surface normal estimation:}
We also evaluate the learned spatial representation to predict surface normals from a single image, another fundamental mid-level vision task that requires spatial understanding of the geometry of the surfaces~\cite{fouhey2013data}. We adopt the the state-of-the-art pyramid scene parsing network PSPNet architecture~\cite{zhao2017pyramid} for surface normal prediction, again swapping in our pre-trained \name~network for the RGB feature backbone.

\vspace*{-0.05in}
\paragraph{Visual navigation:}
Finally, we validate on an embodied visual navigation task. In this task, the agent receives a sequence of RGB images as input and a point goal defined by a displacement vector relative to the starting position of the agent~\cite{anderson2018evaluation}. The agent is spawned at random locations and must navigate to the target location quickly and accurately. This entails reasoning about 3D spatial configurations to avoid obstacles and find the shortest path. We adopt a state-of-the-art reinforcement learning-based PointGoal visual navigation model~\cite{savva_habitat:_2019}.  It consists of a three-layer convolutional network and a fully-connected layer to extract visual feature from the RGB images. We pre-train its visual network using \name, then train the full network end to end.

\vspace*{0.03in}
While other architectures are certainly possible for each task, our choices are based on both on the methods' effectiveness in practice, their wide use in the literature, and code availability.  Our contribution is feature learning from echoes as a pre-training mechanism for spatial tasks, which is orthogonal to advances on architectures for each individual task.  In fact, a key message of our results is that the \name-Net encoder boosts multiple spatial tasks, under multiple different architectures, and on multiple datasets.

\vspace*{-0.1in}
\section{Experiments}~\label{sec:results}
\vspace*{-0.15in}

We present experiments to validate \name~for three tasks and three datasets (Replica~\cite{straub2019replica}, NYU-V2~\cite{silberman2012indoor}, and DIODE~\cite{diode_dataset}). The goal is to examine the impact of our features compared to either learning features for that task from scratch or learning features with  manual semantic supervision. See Supp. for details of the three datasets.

\vspace*{-0.05in}
\paragraph{\textbf{Implementation Details:}}
All networks are implemented in PyTorch.  For the echoes, we use the first 60 ms, which allows most of the room echo responses following the 3 ms chirp to be received. We use an audio sampling rate of 44.1 kHz.  STFT is computed using a Hann window of length 64, hop length of 16, and FFT size of 512. The audio-visual fusion layer (see Fig.~\ref{fig:network}) concatenates the visual and audio feature, and then uses a fully-connected layer to reduce the feature dimension to $D = 128$. See Supp.~for details of the network architectures and optimization hyperparameters.

\vspace*{-0.05in}
\paragraph{\textbf{Evaluation Metrics:}} We report  standard metrics for the downstream tasks.
1) \emph{Monocular Depth Prediction:} RMS, REL, and others as defined above, following~\cite{hu2019revisiting,eigen2014depth}.
2) \emph{Surface Normal Estimation:}
 mean and median of the angle distance and the percentage of good pixels (i.e., the fraction of pixels with cosine distance to ground-truth less than $t$) with
$t = 11.25^{\circ}, 22.5^{\circ}, 30^{\circ}$, following~\cite{fouhey2013data}.
3) \emph{Visual Navigation:}
 success rate normalized by inverse path length (SPL), the distance to the goal at the
end of the episode, and the distance to the goal normalized by the trajectory length, following~\cite{anderson2018evaluation}.

\vspace*{-0.15in}
\subsection{Transferring \name~Features for \textsc{RGB2Depth}}~\label{sec:rgb2depth}
\vspace*{-0.2in}

Having confirmed echoes reveal spatial cues in Sec.~\ref{sec:case-study},
we now examine the effectiveness of \name, our learned representation. Our model achieves $66\%$ test accuracy on the orientation prediction pretext task, while chance performance is only $25\%$; this shows learning the visual-echo consistency task itself is possible.

\begin{table}[t]
\fontsize{7}{10} \selectfont
\centering
\begin{subtable}{\linewidth}\centering
{\begin{tabular}{@{}cc?{0.5mm}*{5}{c|}c}
    \multicolumn{1}{c}{} & & RMS~$\downarrow$ & REL~$\downarrow$ & $~~\log10~~\downarrow$ & $\delta<1.25~\uparrow$ &  $\delta<1.25^2~\uparrow$ &  $\delta<1.25^3~\uparrow$     \\ \specialrule{.12em}{.1em}{.1em}
 	\multirow{2}*{\rotatebox{90}{Sup}}
    & ImageNet Pre-trained  &   0.356   &  0.203   &    0.076    &    0.748            &    0.891            &     0.948            \\
    &  MIT Indoor Scene Pre-trained  &  0.334  &  0.196   &    0.072   & 0.770               &    0.897            &    0.950           \\ \cline{2-8}
	\multirow{2}*{\rotatebox{90}{Unsup}}
    & Scratch &  0.360   &  0.214   &    0.078   &    0.747            &       0.879         &   0.940            \\
    & \name~(Ours)  &  \textbf{0.332}    &   \textbf{0.195}  &    \textbf{0.070}    &   \textbf{0.773}             &    \textbf{0.899}             &    \textbf{0.951}            \\
   \specialrule{.12em}{.1em}{.1em}
    \end{tabular}
}
\caption{Replica}
\end{subtable}
\begin{subtable}{\linewidth}\centering
{\begin{tabular}{@{}cc?{0.5mm}*{5}{c|}c}
    \multicolumn{1}{c}{} & & RMS~$\downarrow$ & REL~$\downarrow$ & $~~\log10~~\downarrow$ & $\delta<1.25~\uparrow$ &  $\delta<1.25^2~\uparrow$ &  $\delta<1.25^3~\uparrow$     \\ \specialrule{.12em}{.1em}{.1em}
 	\multirow{2}*{\rotatebox{90}{Sup}}
    & ImageNet Pre-trained &   0.812   &  0.249   &    0.102    &    0.589            &    0.855            &     0.955           \\
    &  MIT Indoor Scene Pre-trained  &  0.776  &  0.239  &    0.098   &    0.610            &    0.869            &    0.959           \\ \cline{2-8}
	\multirow{2}*{\rotatebox{90}{Unsup}}
    & Scratch  &  0.818   &  0.252   &    0.103    &    0.586            &       0.853         &   0.950           \\
    & \name~(Ours)  &    \textbf{0.797}    &   \textbf{0.246}  &    \textbf{0.100}    &   \textbf{0.600}             &    \textbf{0.863}             &    \textbf{0.956}  \\
   \specialrule{.12em}{.1em}{.1em}
    \end{tabular}
}
\caption{NYU-V2}
\end{subtable}
\begin{subtable}{\linewidth}\centering
{\begin{tabular}{@{}cc?{0.5mm}*{5}{c|}c}
    \multicolumn{1}{c}{} & & RMS~$\downarrow$ & REL~$\downarrow$ & $~~\log10~~\downarrow$ & $\delta<1.25~\uparrow$ &  $\delta<1.25^2~\uparrow$ &  $\delta<1.25^3~\uparrow$     \\ \specialrule{.12em}{.1em}{.1em}
 	\multirow{2}*{\rotatebox{90}{Sup}}
    & ImageNet Pre-trained &   2.250   &  0.453   &    0.199      &   0.336              &    0.591              &     0.766             \\
    &  MIT Indoor Scene Pre-trained  &   2.218   &  0.424   &    0.198      &   0.363              &    0.632              &     0.776            \\ \cline{2-8}
	\multirow{2}*{\rotatebox{90}{Unsup}}
    & Scratch  &   2.352   &  0.481   &    0.214      &   0.321              &    0.581              &     0.742           \\
    & \name~(Ours)  &   \textbf{2.223}   &  \textbf{0.430}   &    \textbf{0.198}      &   \textbf{0.340}              &    \textbf{0.610}              &     \textbf{0.769}   \\
   \specialrule{.12em}{.1em}{.1em}
    \end{tabular}
}
\caption{DIODE}
\end{subtable}
\caption{Depth prediction results on the Replica, NYU-V2, and DIODE datasets.  We use the \textsc{RGB2Depth} network from Sec.~\ref{sec:case-study} for all models. Our \name~pre-training transfers well, consistently predicting depth better than the model trained from scratch.  Furthermore, it is even competitive with the supervised models, whether they are pre-trained for ImageNet or MIT Indoor Scenes (1M/16K manually labeled images). $\downarrow$ lower better, $\uparrow$ higher better. (Un)sup = (un)supervised. We boldface the best unsupervised method.}
\label{table:unet_exp}
\vspace*{-0.1in}
\end{table}

First, we use the same \textsc{RGB2Depth} network from our case study in Sec.~\ref{sec:case-study} as a testbed to demonstrate the learned spatial features can be successfully transferred to other domains. Instead of randomly initializing the \textsc{RGB2Depth} UNet encoder, we initialize with an encoder 1) pre-trained for our visual-echo consistency task, 2) pre-trained for image classification using ImageNet~\cite{deng2009imagenet}, or 3) pre-trained for scene classification using the MIT Indoor Scene dataset~\cite{quattoni2009recognizing}. Throughout, aside from the standard ImageNet pre-training baseline, we also include MIT Indoor Scenes pre-training, in case it strengthens the baseline due to its domain alignment with the indoor scenes in Replica, DIODE, and NYU-2.\footnote{Like the test datasets, MIT Indoor Scenes contains indoor scenes.  Performance is similar when pre-training on Places~\cite{zhou2017places}, which is larger but contains diverse indoor and outdoor scenes.}

Table~\ref{table:unet_exp} shows the results on all three datasets: Replica, NYU-V2, and DIODE. The model initialized with our pre-trained \name~network achieves much better performance compared to the model trained from scratch. Moreover, it even outperforms the supervised model pre-trained on scene classification in some cases. The ImageNet pre-trained model performs much worse; we suspect that the UNet encoder does not have sufficient capacity to handle ImageNet classification, and also the ImageNet domain is much different than indoor scene environments. This result accentuates that task similarity promotes positive transfer~\cite{Zamir_2018_CVPR}: our unsupervised spatial pre-training task is more powerful for depth inference than a supervised semantic category pre-training task. See Supp.~for low-shot experiments varying the amount of training data. 

We also perform an ablation study to demonstrate that the design of our spatial representation learning framework is essential and effective. We compare with the following two variants: \textsc{SimpleVisualEchoes}, which simplifies our orientation prediction task to two classes; and BinaryMatching, which mimics prior work~\cite{arandjelovic2017look} that leverages the correspondence between images and audio as supervision by training a network to decide if the echo and RGB are from the same environment. As shown in Table~\ref{tab:ablation_study}, our method performs much better than both baselines. See Supp. for details.

\begin{table*}[t]
\fontsize{7}{10} \selectfont
    \centering
\begin{tabular}{zc|c|c|c|c|c}
\toprule
              & ~~RMS~~~$\downarrow$ & ~~REL~~~$\downarrow$ & $~~\log10~~\downarrow$ & $\delta<1.25~\uparrow$ &  $\delta<1.25^2~\uparrow$ &  $\delta<1.25^3~\uparrow$ \\ \hline
\textsc{Scratch}       &  0.360   &  0.214   &    0.078   &    0.747            &       0.879         &   0.940            \\ 
\textsc{SimpleVisualEchoes}    &   0.340   &  0.198   &    0.073    &    0.763            &    0.892            &     0.948           \\ 
\textsc{BinaryMatching}     &  0.345    &   0.199  &    0.074    &   0.760             &    0.889             &    0.944             \\
\textsc{VisualEchoes~(Ours)} &  \textbf{0.332}    &   \textbf{0.195}  &    \textbf{0.070}    &   \textbf{0.773}             &    \textbf{0.899}             &    \textbf{0.951}             \\ \bottomrule
\end{tabular}
\caption{Ablation study on Replica. See Supp. for results on NYU-V2 and Diode.}
\vspace{-0.05in}
\label{tab:ablation_study}
\end{table*}

\vspace*{-0.1in}
\subsection{Evaluating on Downstream Tasks}~\label{sec:downstream_tasks_exp}
\vspace*{-0.2in}

Next we evaluate the impact of our learned \name~representation on all three downstream tasks introduced in Sec.~\ref{sec:downstream_tasks}.

\begin{table*}
	\fontsize{7}{10} \selectfont
    \centering
\begin{subtable}{\linewidth}\centering
{\begin{tabular}{@{}cc?{0.5mm}*{5}{c|}c}
    \multicolumn{1}{c}{} & & RMS~$\downarrow$ & REL~$\downarrow$ & $~~\log10~~\downarrow$ & $\delta<1.25~\uparrow$ &  $\delta<1.25^2~\uparrow$ &  $\delta<1.25^3~\uparrow$     \\ \specialrule{.12em}{.1em}{.1em}
 	\multirow{2}*{\rotatebox{90}{Sup}}
    & ImageNet Pre-trained~\cite{hu2019revisiting} &  \textbf{0.555}   &  \textbf{0.126}   &  \textbf{0.054}    &   \textbf{0.843}             &      \textbf{0.968}          &    \textbf{0.991}            \\
    &  MIT Indoor Scene Pre-trained  & 0.711   &  0.180   &    0.075   &   0.730             &     0.925          &    0.979           \\ \cline{2-8}
	\multirow{2}*{\rotatebox{90}{Unsup}}
    & Scratch &  0.804   &  0.209   &    0.086    &   0.676             &      0.897          &    0.967            \\
    & \name~(Ours)  &  0.683   &  0.165   &    0.069   &   0.762             &      0.934          &    0.981            \\
   \specialrule{.12em}{.1em}{.1em}
    \end{tabular}
    \caption{Depth prediction results on NYU-V2. }
    \label{table:depth_prediction}
}
\end{subtable}
\begin{subtable}{\linewidth}\centering
{\begin{tabular}{@{}cc?{0.5mm}*{4}{c|}c}
    \multicolumn{1}{c}{} & & Mean Dist.~$\downarrow$ & Median Dist.~$\downarrow$  & $t < 11.25^{\circ}$~$\uparrow$ & $t < 22.5^{\circ}$~$\uparrow$ &  $t < 30^{\circ}$~$\uparrow$     \\ \specialrule{.12em}{.1em}{.1em}
 	\multirow{2}*{\rotatebox{90}{Sup}}
    & ImageNet Pre-trained       &  26.4   &  17.1   &    36.1    &    59.2            &       68.5                \\ 
    & MIT Indoor Scene Pre-trained        &  25.2   &  17.5   &    36.5    &    57.8            &       67.2                 \\  \cline{2-7}
	\multirow{2}*{\rotatebox{90}{Unsup}}
    & Scratch     &  26.3    &   16.1  &    37.9    &   60.6             &    69.0                        \\
    & \name~(Ours)  &  \textbf{22.9}   &  \textbf{14.1}  &   \textbf{42.7}    &   \textbf{64.1}             &      \textbf{72.4}                    \\    \specialrule{.12em}{.1em}{.1em}
    \end{tabular}
    \caption{Surface normal estimation results on NYU-V2. The results for the ImageNet Pre-trained baseline and the Scratch baseline are directly quoted from~\cite{goyal2019scaling}.}
    \label{tabel:surface_normal}
}
\end{subtable} 
\begin{subtable}{\linewidth}\centering
{\begin{tabular}{@{}cc?{0.5mm}*{2}{c|}c}
    \multicolumn{1}{c}{} & & ~~~SPL~$\uparrow$~~~ & Distance to Goal~$\downarrow$ & Normalized Distance to Goal~$\downarrow$     \\ \specialrule{.12em}{.1em}{.1em}
 	\multirow{2}*{\rotatebox{90}{Sup}}
    & ImageNet Pre-trained   &   0.833  &  0.663  &    0.081   \\ 
    & MIT Indoor Scene Pre-trained   &   0.798  &  1.05  &    0.124   \\ \cline{2-5}
	\multirow{2}*{\rotatebox{90}{Unsup}}
    & Scratch   &   0.830  &  0.728  &    0.096   \\ 
    & \name~(Ours)  &  \textbf{0.856}   &  \textbf{0.476}   &    \textbf{0.061}   \\ 
 \specialrule{.12em}{.1em}{.1em}
    \end{tabular}
\caption{Visual navigation performance in unseen Replica environments.}
\label{table:navigation}
}
\end{subtable} 
\caption{Results for three downstream tasks. $\downarrow$ lower better, $\uparrow$ higher better.}
\label{table:downstream_tasks}
\vspace*{-0.1in}
\end{table*}

\vspace*{-0.2in}
\subsubsection{Monocular depth prediction:}
Table~\ref{table:depth_prediction} shows the results.\footnote{We evaluate on NYU-V2, the most widely used dataset for the task of single view depth prediction and surface normal estimation. The authors's code~\cite{hu2019revisiting,goyal2019scaling} is tailored to this dataset.} All methods use the same settings as~\cite{hu2019revisiting}, where they evaluate and report results on NYU-V2. We use the authors' publicly available code\footnote{\url{https://github.com/JunjH/Revisiting_Single_Depth_Estimation}} and use ResNet-50 as the encoder.  See Supp.~for details. With this apples-to-apples comparison, the difference in performance can be attributed to whether/how the encoder is pre-trained. Although our \name~features are learned from Replica, they transfer reasonably well to NYU-V2, outperforming models trained from scratch by a large margin. This result is important because it shows that despite training with simulated audio, our model generalizes to real-world test images. Our features also compare favorably to supervised models trained with heavy supervision. 

\vspace*{-0.2in}
\subsubsection{Surface normal estimation:} Table~\ref{tabel:surface_normal} shows the results. We follow the same setting as~\cite{goyal2019scaling} and we use the authors' publicly available code.\footnote{\url{https://github.com/facebookresearch/fair_self_supervision_benchmark}} Our model performs much better even compared to the ImageNet-supervised pre-trained model, demonstrating that our interaction-based feature learning framework via echoes makes the learned features more useful for 3D geometric tasks.

\vspace*{-0.05in}
\paragraph{Visual navigation:} Table~\ref{table:navigation} shows the results. By pre-training the visual network, \name~equips the embodied agents with a better sense of room geometry and allows them to learn faster (see Supp. for training curves). Notably, the agent also ends much closer to the goal.  We suspect it can better gauge the distance because of our \name~pre-training. Models pre-trained for classification on MIT Indoor Scenes perform more poorly than Scratch; again, this suggests features useful for recognition may not be optimal for a spatial task like point goal navigation.

This series of results on three tasks consistently shows the promise of our \name~features. We see that learning from echoes translates into a strengthened \emph{visual} encoding. Importantly, while it is always an option to train multiple representations entirely from scratch to support each given task, our results are encouraging since they show the \emph{same} fundamental interaction-based pre-training is versatile across multiple tasks.

\vspace*{-0.1in}
\subsection{Qualitative Results}
\begin{figure}[t]
    \center
    \includegraphics[width=\linewidth]{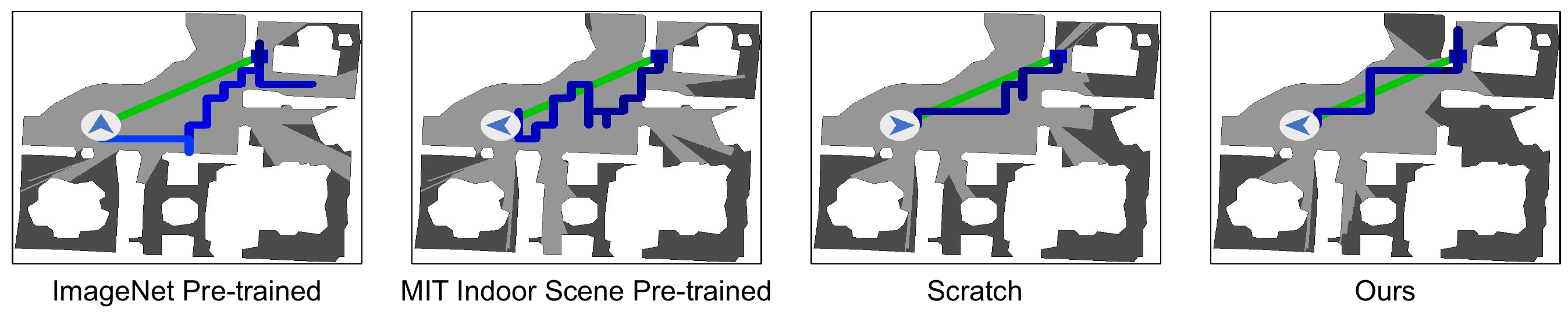}
    \caption{Qualitative examples of visual navigation trajectories on top-down maps. Blue square and arrow denote agent’s starting and ending positions, respectively. The green path indicates the shortest geodesic path to the goal, and the agent's path is in dark blue. Agent path color fades from dark blue to light blue as time goes by. Note, the agent sees a sequence of egocentric views, not the map.}
    \label{fig:navigation_qualitative}
\vspace*{-0.1in}
\end{figure}

Fig.~\ref{fig:navigation_qualitative} shows example navigation trajectories on top-down maps. Our visual-echo consistency pre-training task allows the agent to better interpret the room's spatial layout to find the goal more quickly than the baselines. See Supp.~for qualitative results on depth estimation and surface normal examples. Initializing with our pre-trained \name~network leads to much more accurate depth prediction and surface normal estimates compared to no pre-training, demonstrating the usefulness of the learned spatial features.

\vspace{-0.1in}
\section{Conclusions and Future Work}
\vspace*{-0.05in}

We presented an approach to learn spatial image representations via echolocation. We performed an in-depth study on the spatial cues contained in echoes and how they can inform single-view depth estimation. We showed that the learned spatial features can benefit three downstream vision tasks. Our work opens a new path for interaction-based representation learning for embodied agents and demonstrates the potential of learning spatial visual representations even with a limited amount of multisensory data. 

While our current implementation learns from audio rendered in a simulator, the results  show that the learned spatial features already benefit transfer to vision-only tasks in real photos outside of the scanned environments (e.g., the NYU-V2~\cite{silberman2012indoor} and DIODE~\cite{diode_dataset} images), indicating the realism of what our system learned.  Nonetheless, it will be interesting future work to capture the echoes on a real robot.  We are also interested in pursuing these ideas within a sequential model, such that the agent could actively decide when to emit chirps and what type of chirps to emit to get the most informative echo responses.


\vspace{-0.05in}
\paragraph{Acknowledgements:} UT Austin is supported in part by DARPA Lifelong Learning Machines and ONR PECASE. RG is supported by Google PhD Fellowship and Adobe Research Fellowship.\\

\noindent Supplementary materials: \\
\url{http://vision.cs.utexas.edu/projects/visualEchoes/supplementary.pdf}

\clearpage
%
%
\bibliographystyle{splncs04}
\bibliography{ref_RG}

\begin{thebibliography}{10}
\providecommand{\url}[1]{\texttt{#1}}
\providecommand{\urlprefix}{URL }
\providecommand{\doi}[1]{https://doi.org/#1}

\bibitem{agrawal2015learning}
Agrawal, P., Carreira, J., Malik, J.: Learning to see by moving. In: ICCV
  (2015)

\bibitem{agrawal2016learning}
Agrawal, P., Nair, A.V., Abbeel, P., Malik, J., Levine, S.: Learning to poke by
  poking: Experiential learning of intuitive physics. In: NeurIPS (2016)

\bibitem{alameda2015salsa}
Alameda-Pineda, X., Staiano, J., Subramanian, R., Batrinca, L., Ricci, E.,
  Lepri, B., Lanz, O., Sebe, N.: Salsa: A novel dataset for multimodal group
  behavior analysis. TPAMI  (2015)

\bibitem{anderson2018evaluation}
Anderson, P., Chang, A., Chaplot, D.S., Dosovitskiy, A., Gupta, S., Koltun, V.,
  Kosecka, J., Malik, J., Mottaghi, R., Savva, M., et~al.: On evaluation of
  embodied navigation agents. arXiv preprint arXiv:1807.06757  (2018)

\bibitem{antonacci2012inference}
Antonacci, F., Filos, J., Thomas, M.R., Habets, E.A., Sarti, A., Naylor, P.A.,
  Tubaro, S.: Inference of room geometry from acoustic impulse responses. IEEE
  Transactions on Audio, Speech, and Language Processing  (2012)

\bibitem{arandjelovic2017look}
Arandjelovic, R., Zisserman, A.: Look, listen and learn. In: ICCV (2017)

\bibitem{arandjelovic2017objects}
Arandjelovi{\'c}, R., Zisserman, A.: Objects that sound. In: ECCV (2018)

\bibitem{aytar2016soundnet}
Aytar, Y., Vondrick, C., Torralba, A.: Soundnet: Learning sound representations
  from unlabeled video. In: NeurIPS (2016)

\bibitem{ban2018icassp}
Ban, Y., Li, X., Alameda-Pineda, X., Girin, L., Horaud, R.: Accounting for room
  acoustics in audio-visual multi-speaker tracking. In: ICASSP (2018)

\bibitem{chang_matterport3d_2017}
Chang, A., Dai, A., Funkhouser, T., Halber, M., Niessner, M., Savva, M., Song,
  S., Zeng, A., Zhang, Y.: Matterport3d: Learning from rgb-d data in indoor
  environments. 3DV  (2017)

\bibitem{changan-eccv2020}
Chen, C., Jain, U., Schissler, C., Gari, S.V.A., Al-Halah, Z., Ithapu, V.K.,
  Robinson, P., Grauman, K.: Audio-visual embodied navigation. In: ECCV (2020)

\bibitem{batvision}
Christensen, J., Hornauer, S., Yu, S.: Batvision - learning to see 3d spatial
  layout with two ears. In: ICRA (2020)

\bibitem{deng2009imagenet}
Deng, J., Dong, W., Socher, R., Li, L.J., Li, K., Fei-Fei, L.: Imagenet: A
  large-scale hierarchical image database. In: CVPR (2009)

\bibitem{dokmanic2013acoustic}
Dokmani{\'c}, I., Parhizkar, R., Walther, A., Lu, Y.M., Vetterli, M.: Acoustic
  echoes reveal room shape. Proceedings of the National Academy of Sciences
  (2013)

\bibitem{eigen2015predicting}
Eigen, D., Fergus, R.: Predicting depth, surface normals and semantic labels
  with a common multi-scale convolutional architecture. In: ICCV (2015)

\bibitem{eigen2014depth}
Eigen, D., Puhrsch, C., Fergus, R.: Depth map prediction from a single image
  using a multi-scale deep network. In: NeurIPS (2014)

\bibitem{eliakim2018fully}
Eliakim, I., Cohen, Z., Kosa, G., Yovel, Y.: A fully autonomous terrestrial
  bat-like acoustic robot. PLoS computational biology  (2018)

\bibitem{ephrat2018looking}
Ephrat, A., Mosseri, I., Lang, O., Dekel, T., Wilson, K., Hassidim, A.,
  Freeman, W.T., Rubinstein, M.: Looking to listen at the cocktail party: A
  speaker-independent audio-visual model for speech separation. In: SIGGRAPH
  (2018)

\bibitem{feng2019self}
Feng, Z., Xu, C., Tao, D.: Self-supervised representation learning by rotation
  feature decoupling. In: CVPR (2019)

\bibitem{fernando2017self}
Fernando, B., Bilen, H., Gavves, E., Gould, S.: Self-supervised video
  representation learning with odd-one-out networks. In: CVPR (2017)

\bibitem{fouhey2013data}
Fouhey, D.F., Gupta, A., Hebert, M.: Data-driven 3d primitives for single image
  understanding. In: ICCV (2013)

\bibitem{Frank2020ComparingVT}
Frank, N., Wolf, L., Olshansky, D., Boonman, A., Yovel, Y.: Comparing
  vision-based to sonar-based 3d reconstruction. ICCP  (2020)

\bibitem{fu2018deep}
Fu, H., Gong, M., Wang, C., Batmanghelich, K., Tao, D.: Deep ordinal regression
  network for monocular depth estimation. In: CVPR (2018)

\bibitem{gan2020music}
Gan, C., Huang, D., Zhao, H., Tenenbaum, J.B., Torralba, A.: Music gesture for
  visual sound separation. In: CVPR (2020)

\bibitem{gan2020look}
Gan, C., Zhang, Y., Wu, J., Gong, B., Tenenbaum, J.B.: Look, listen, and act:
  Towards audio-visual embodied navigation. In: ICRA (2020)

\bibitem{gan2019tracking}
Gan, C., Zhao, H., Chen, P., Cox, D., Torralba, A.: Self-supervised moving
  vehicle tracking with stereo sound. In: ICCV (2019)

\bibitem{gandhi2017learning}
Gandhi, D., Pinto, L., Gupta, A.: Learning to fly by crashing. In: IROS (2017)

\bibitem{gao2018objectSounds}
Gao, R., Feris, R., Grauman, K.: Learning to separate object sounds by watching
  unlabeled video. In: ECCV (2018)

\bibitem{gao2019visualsound}
Gao, R., Grauman, K.: 2.5d visual sound. In: CVPR (2019)

\bibitem{gao2019coseparation}
Gao, R., Grauman, K.: Co-separating sounds of visual objects. In: ICCV (2019)

\bibitem{gao2016object-centric}
Gao, R., Jayaraman, D., Grauman, K.: Object-centric representation learning
  from unlabeled videos. In: ACCV (2016)

\bibitem{gao2020listentolook}
Gao, R., Oh, T.H., Grauman, K., Torresani, L.: Listen to look: Action
  recognition by previewing audio. In: CVPR (2020)

\bibitem{garg2016unsupervised}
Garg, R., BG, V.K., Carneiro, G., Reid, I.: Unsupervised cnn for single view
  depth estimation: Geometry to the rescue. In: ECCV (2016)

\bibitem{gebru2015iccvw}
Gebru, I.D., Ba, S., Evangelidis, G., Horaud, R.: Tracking the active speaker
  based on a joint audio-visual observation model. In: ICCV Workshops (2015)

\bibitem{gidaris2018unsupervised}
Gidaris, S., Singh, P., Komodakis, N.: Unsupervised representation learning by
  predicting image rotations. In: ICLR (2018)

\bibitem{godard2017unsupervised}
Godard, C., Mac~Aodha, O., Brostow, G.J.: Unsupervised monocular depth
  estimation with left-right consistency. In: CVPR (2017)

\bibitem{godard2019digging}
Godard, C., Mac~Aodha, O., Firman, M., Brostow, G.J.: Digging into
  self-supervised monocular depth estimation. In: ICCV (2019)

\bibitem{goyal2019scaling}
Goyal, P., Mahajan, D., Gupta, A., Misra, I.: Scaling and benchmarking
  self-supervised visual representation learning. In: ICCV (2019)

\bibitem{he2016deep}
He, K., Zhang, X., Ren, S., Sun, J.: Deep residual learning for image
  recognition. In: CVPR (2016)

\bibitem{hershey2000audio}
Hershey, J.R., Movellan, J.R.: Audio vision: Using audio-visual synchrony to
  locate sounds. In: NeurIPS (2000)

\bibitem{hu2019revisiting}
Hu, J., Ozay, M., Zhang, Y., Okatani, T.: Revisiting single image depth
  estimation: Toward higher resolution maps with accurate object boundaries.
  In: WACV (2019)

\bibitem{irie2019seeing}
Irie, G., Ostrek, M., Wang, H., Kameoka, H., Kimura, A., Kawanishi, T.,
  Kashino, K.: Seeing through sounds: Predicting visual semantic segmentation
  results from multichannel audio signals. In: ICASSP (2019)

\bibitem{slow-steady}
Jayaraman, D., Grauman, K.: Slow and steady feature analysis: Higher order
  temporal coherence in video. In: CVPR (2016)

\bibitem{jayaraman2018shapecodes}
Jayaraman, D., Gao, R., Grauman, K.: Shapecodes: self-supervised feature
  learning by lifting views to viewgrids. In: ECCV (2018)

\bibitem{jayaraman2015learning}
Jayaraman, D., Grauman, K.: Learning image representations equivariant to
  ego-motion. In: ICCV (2015)

\bibitem{jiang2018self}
Jiang, H., Larsson, G., Maire Greg~Shakhnarovich, M., Learned-Miller, E.:
  Self-supervised relative depth learning for urban scene understanding. In:
  ECCV (2018)

\bibitem{karsch2014depth}
Karsch, K., Liu, C., Kang, S.B.: Depth transfer: Depth extraction from video
  using non-parametric sampling. TPAMI  (2014)

\bibitem{kazakos2019TBN}
Kazakos, E., Nagrani, A., Zisserman, A., Damen, D.: Epic-fusion: Audio-visual
  temporal binding for egocentric action recognition. In: ICCV (2019)

\bibitem{kim20173d}
Kim, H., Remaggi, L., Jackson, P.J., Fazi, F.M., Hilton, A.: 3d room geometry
  reconstruction using audio-visual sensors. In: 3DV (2017)

\bibitem{Korbar2018cotraining}
Korbar, B., Tran, D., Torresani, L.: Co-training of audio and video
  representations from self-supervised temporal synchronization. In: NeurIPS
  (2018)

\bibitem{room_acoustics_taylor}
Kuttruff, H.: Room Acoustics. CRC Press (2017)

\bibitem{larsson2017colorization}
Larsson, G., Maire, M., Shakhnarovich, G.: Colorization as a proxy task for
  visual understanding. In: CVPR (2017)

\bibitem{liu2015deep}
Liu, F., Shen, C., Lin, G.: Deep convolutional neural fields for depth
  estimation from a single image. In: CVPR (2015)

\bibitem{mcguire2019minimal}
McGuire, K., De~Wagter, C., Tuyls, K., Kappen, H., de~Croon, G.: Minimal
  navigation solution for a swarm of tiny flying robots to explore an unknown
  environment. Science Robotics  (2019)

\bibitem{misra2016shuffle}
Misra, I., Zitnick, C.L., Hebert, M.: Shuffle and learn: unsupervised learning
  using temporal order verification. In: ECCV (2016)

\bibitem{morgadoNIPS18}
Morgado, P., Vasconcelos, N., Langlois, T., Wang, O.: Self-supervised
  generation of spatial audio for 360${}^\circ$ video. In: NeurIPS (2018)

\bibitem{noroozi2016unsupervised}
Noroozi, M., Favaro, P.: Unsupervised learning of visual representations by
  solving jigsaw puzzles. In: ECCV (2016)

\bibitem{owens2018audio}
Owens, A., Efros, A.A.: Audio-visual scene analysis with self-supervised
  multisensory features. In: ECCV (2018)

\bibitem{owens2016visually}
Owens, A., Isola, P., McDermott, J., Torralba, A., Adelson, E.H., Freeman,
  W.T.: Visually indicated sounds. In: CVPR (2016)

\bibitem{owens2016ambient}
Owens, A., Wu, J., McDermott, J.H., Freeman, W.T., Torralba, A.: Ambient sound
  provides supervision for visual learning. In: ECCV (2016)

\bibitem{palossi201964}
Palossi, D., Loquercio, A., Conti, F., Flamand, E., Scaramuzza, D., Benini, L.:
  A 64-mw dnn-based visual navigation engine for autonomous nano-drones. IEEE
  Internet of Things Journal  (2019)

\bibitem{pinto-icra2016}
Pinto, L., Gupta, A.: Supersizing self-supervision: Learning to grasp from 50k
  tries and 700 robot hours. In: ICRA (2016)

\bibitem{purushwalkam2019bounce}
Purushwalkam, S., Gupta, A., Kaufman, D.M., Russell, B.: Bounce and learn:
  Modeling scene dynamics with real-world bounces. In: ICLR (2019)

\bibitem{quattoni2009recognizing}
Quattoni, A., Torralba, A.: Recognizing indoor scenes. In: CVPR (2009)

\bibitem{Ranjan_2019_CVPR}
Ranjan, A., Jampani, V., Balles, L., Kim, K., Sun, D., Wulff, J., Black, M.J.:
  Competitive collaboration: Joint unsupervised learning of depth, camera
  motion, optical flow and motion segmentation. In: CVPR (2019)

\bibitem{ren2018cross}
Ren, Z., Jae~Lee, Y.: Cross-domain self-supervised multi-task feature learning
  using synthetic imagery. In: CVPR (2018)

\bibitem{ronneberger2015u}
Ronneberger, O., Fischer, P., Brox, T.: U-net: Convolutional networks for
  biomedical image segmentation. In: MICCAI (2015)

\bibitem{rosenblum2000echolocating}
Rosenblum, L.D., Gordon, M.S., Jarquin, L.: Echolocating distance by moving and
  stationary listeners. Ecological Psychology  (2000)

\bibitem{de1994learning}
de~Sa, V.R.: Learning classification with unlabeled data. In: NeurIPS (1994)

\bibitem{savva_habitat:_2019}
Savva, M., Kadian, A., Maksymets, O., Zhao, Y., Wijmans, E., Jain, B., Straub,
  J., Liu, J., Koltun, V., Malik, J., Parikh, D., Batra, D.: Habitat: {A}
  {P}latform for {E}mbodied {AI} {R}esearch. In: ICCV (2019)

\bibitem{Senocak_2018_CVPR}
Senocak, A., Oh, T.H., Kim, J., Yang, M.H., So~Kweon, I.: Learning to localize
  sound source in visual scenes. In: CVPR (2018)

\bibitem{silberman2012indoor}
Silberman, N., Hoiem, D., Kohli, P., Fergus, R.: Indoor segmentation and
  support inference from rgbd images. In: ECCV (2012)

\bibitem{straub2019replica}
Straub, J., Whelan, T., Ma, L., Chen, Y., Wijmans, E., Green, S., Engel, J.J.,
  Mur-Artal, R., Ren, C., Verma, S., et~al.: The replica dataset: A digital
  replica of indoor spaces. arXiv preprint arXiv:1906.05797  (2019)

\bibitem{stroffregen1995human}
Stroffregen, T.A., Pittenger, J.B.: Human echolocation as a basic form of
  perception and action. Ecological psychology  (1995)

\bibitem{tian2018audio}
Tian, Y., Shi, J., Li, B., Duan, Z., Xu, C.: Audio-visual event localization in
  unconstrained videos. In: ECCV (2018)

\bibitem{Ummenhofer_2017_CVPR}
Ummenhofer, B., Zhou, H., Uhrig, J., Mayer, N., Ilg, E., Dosovitskiy, A., Brox,
  T.: Demon: Depth and motion network for learning monocular stereo. In: CVPR
  (2017)

\bibitem{vanderelst2015sensorimotor}
Vanderelst, D., Holderied, M.W., Peremans, H.: Sensorimotor model of obstacle
  avoidance in echolocating bats. PLoS computational biology  (2015)

\bibitem{diode_dataset}
Vasiljevic, I., Kolkin, N., Zhang, S., Luo, R., Wang, H., Dai, F.Z., Daniele,
  A.F., Mostajabi, M., Basart, S., Walter, M.R., Shakhnarovich, G.: {DIODE}:
  {A} {D}ense {I}ndoor and {O}utdoor {DE}pth {D}ataset. arXiv preprint
  arXiv:1908.00463  (2019)

\bibitem{veach1995bidirectional}
Veach, E., Guibas, L.: Bidirectional estimators for light transport. In:
  Photorealistic Rendering Techniques (1995)

\bibitem{vijayanarasimhan2017sfm}
Vijayanarasimhan, S., Ricco, S., Schmid, C., Sukthankar, R., Fragkiadaki, K.:
  Sfm-net: Learning of structure and motion from video. arXiv preprint
  arXiv:1704.07804  (2017)

\bibitem{wang2015towards}
Wang, P., Shen, X., Lin, Z., Cohen, S., Price, B., Yuille, A.L.: Towards
  unified depth and semantic prediction from a single image. In: CVPR (2015)

\bibitem{wang2015unsupervised}
Wang, X., Gupta, A.: Unsupervised learning of visual representations using
  videos. In: ICCV (2015)

\bibitem{xia_gibson_2018}
Xia, F., Zamir, A.R., He, Z., Sax, A., Malik, J., Savarese, S.: Gibson env:
  Real-world perception for embodied agents. In: CVPR (2018)

\bibitem{xu2017multi}
Xu, D., Ricci, E., Ouyang, W., Wang, X., Sebe, N.: Multi-scale continuous crfs
  as sequential deep networks for monocular depth estimation. In: CVPR (2017)

\bibitem{yang2018unsupervised}
Yang, Z., Wang, P., Xu, W., Zhao, L., Nevatia, R.: Unsupervised learning of
  geometry with edge-aware depth-normal consistency. In: AAAI (2018)

\bibitem{ye20153d}
Ye, M., Zhang, Y., Yang, R., Manocha, D.: 3d reconstruction in the presence of
  glasses by acoustic and stereo fusion. In: ICCV (2015)

\bibitem{Zamir_2018_CVPR}
Zamir, A.R., Sax, A., Shen, W., Guibas, L.J., Malik, J., Savarese, S.:
  Taskonomy: Disentangling task transfer learning. In: CVPR (2018)

\bibitem{zhang2016colorful}
Zhang, R., Isola, P., Efros, A.A.: Colorful image colorization. In: ECCV (2016)

\bibitem{zhao2018sound}
Zhao, H., Gan, C., Rouditchenko, A., Vondrick, C., McDermott, J., Torralba, A.:
  The sound of pixels. In: ECCV (2018)

\bibitem{zhao2017pyramid}
Zhao, H., Shi, J., Qi, X., Wang, X., Jia, J.: Pyramid scene parsing network.
  In: CVPR (2017)

\bibitem{zhou2017places}
Zhou, B., Lapedriza, A., Khosla, A., Oliva, A., Torralba, A.: Places: A 10
  million image database for scene recognition. TPAMI  (2017)

\bibitem{zhou2017unsupervised}
Zhou, T., Brown, M., Snavely, N., Lowe, D.G.: Unsupervised learning of depth
  and ego-motion from video. In: CVPR (2017)

\bibitem{zhou2017visual}
Zhou, Y., Wang, Z., Fang, C., Bui, T., Berg, T.L.: Visual to sound: Generating
  natural sound for videos in the wild. In: CVPR (2018)

\end{thebibliography}
\end{document}